\documentclass[letterpaper, 10 pt, conference]{ieeeconf}  

\IEEEoverridecommandlockouts                              

\usepackage{xcolor}
\usepackage{amsmath,amssymb,amsfonts,color,nicefrac}
\usepackage{graphicx}
\usepackage{algorithm}
\usepackage{algpseudocode}
\usepackage{comment}

\usepackage{caption}
\usepackage{subcaption}
\usepackage{cite}

\DeclareMathOperator*{\argmin}{arg\,min}

\DeclareSymbolFont{matha}{OML}{txmi}{m}{it}
\DeclareMathSymbol{\varv}{\mathord}{matha}{118}
\renewcommand{\Re}{\mathbb{R}}
\newcommand{\A}{\mathcal{A}}

\title{\LARGE \bf
Communication-Aware Iterative Map Compression \\for Online Path-Planning
}

\author{Evangelos Psomiadis, Ali Reza Pedram, Dipankar Maity, Panagiotis Tsiotras
\thanks{The work was supported by the ARL grant DCIST CRA W911NF-17-2-0181.}
\thanks{E. Psomiadis, A. R. Pedram, and P. Tsiotras are with the D. Guggenheim School of Aerospace Engineering, Georgia Institute of Technology, Atlanta, GA, 30332-0150, USA. Email:
       \{epsomiadis3, apedram3, tsiotras\}@gatech.edu}%
\thanks{D. Maity is with the Department of Electrical and Computer Engineering, University of North Carolina at Charlotte, NC, 28223-0001, USA. Email:
       dmaity@charlotte.edu} %
}

\begin{document}
\maketitle
\thispagestyle{empty}
\pagestyle{empty}

\begin{abstract}
This paper addresses the problem of optimizing communicated information among heterogeneous, resource-aware robot teams to facilitate their navigation.
In such operations, a mobile robot compresses its local map to assist another robot in reaching a target within an uncharted environment. 
The primary challenge lies in ensuring that the map compression step balances network load while transmitting only the most essential information for effective navigation.
We propose a communication framework that sequentially selects the optimal map compression in a task-driven, communication-aware manner. 
It introduces a decoder capable of iterative map estimation, handling noise through Kalman filter techniques. 
The computational speed of our decoder allows for a larger compression template set compared to previous methods, and enables applications in more challenging environments.
Specifically, our simulations demonstrate a remarkable 98\% reduction in communicated information, compared to a framework that transmits the raw data, on a large Mars inclination map and an Earth map, all while maintaining similar planning costs.
Furthermore, our method significantly reduces computational time compared to the state-of-the-art approach.
\end{abstract}

\section{Introduction}

Situational awareness is a key element for the success of collaborative multi-robot missions ranging from search and rescue operations on Earth \cite{Queralta2020} to extraterrestrial space missions \cite{Thangavelautham2020}. 
These missions involve robots collaboratively exploring an unknown region  while exchanging \textit{relevant} information with their teammates to improve team knowledge, thus developing global situational awareness.

Multi-robot collaboration begets several challenges, including map fusion \cite{miller2022stronger} and intermittent communication \cite{schack2024sound}.
Existing approaches, such as \cite{yue2020collaborative} and \cite{cunningham2013ddf}, address map fusion by merging local semantic and geometric maps, respectively, into a unified global representation.
However, in real-world scenarios, communication is not always guaranteed. 
Recent works, such as \cite{schack2024sound}, leverage communication loss as a signal to refine a robot’s belief about its teammate’s state and infer information about unvisited areas.
Similar ideas are also explored in the controls community \cite{maity2020minimal} proving that \textit{no communication can serve as a form of (implicit) communication}.

In multi-robot navigation, path-planning solutions must account for both task requirements and communication constraints. 
Existing approaches often treat communication as an afterthought \cite{chang2023}, reducing their overall effectiveness. 
Thus, it is crucial to develop task-relevant, resource-aware compression schemes that enable the transmission of critical information in a timely manner and at an optimal resolution. 
Determining \textit{which} information to send, \textit{when} to send, and at \textit{what} resolution are the key decision-making factors.

The coupled decision-making and communication problem has also been addressed in the controls community \cite{delchamps1990stabilizing, brockett2000quantized,  nair2004stabilizability, kostina2019rate}.
However, developing the optimal compression/quantization scheme still remains intractable, even for linear systems \cite{yüksel2019note}. 
Recently, there have been studies on how to choose the \textit{optimal} quantizer from a given set in a task- and communication-aware manner \cite{maity2021optimal, maity_quant, afshari2024communication}, alleviating the intractability of the quantizer design problem.
In this work, we follow the same principle, where a set of quantizers is available to the robots to compress their perception data before transmitting it to another teammate. 

\begin{figure}[tb]
    \centering        
    {\includegraphics[width=1\linewidth]{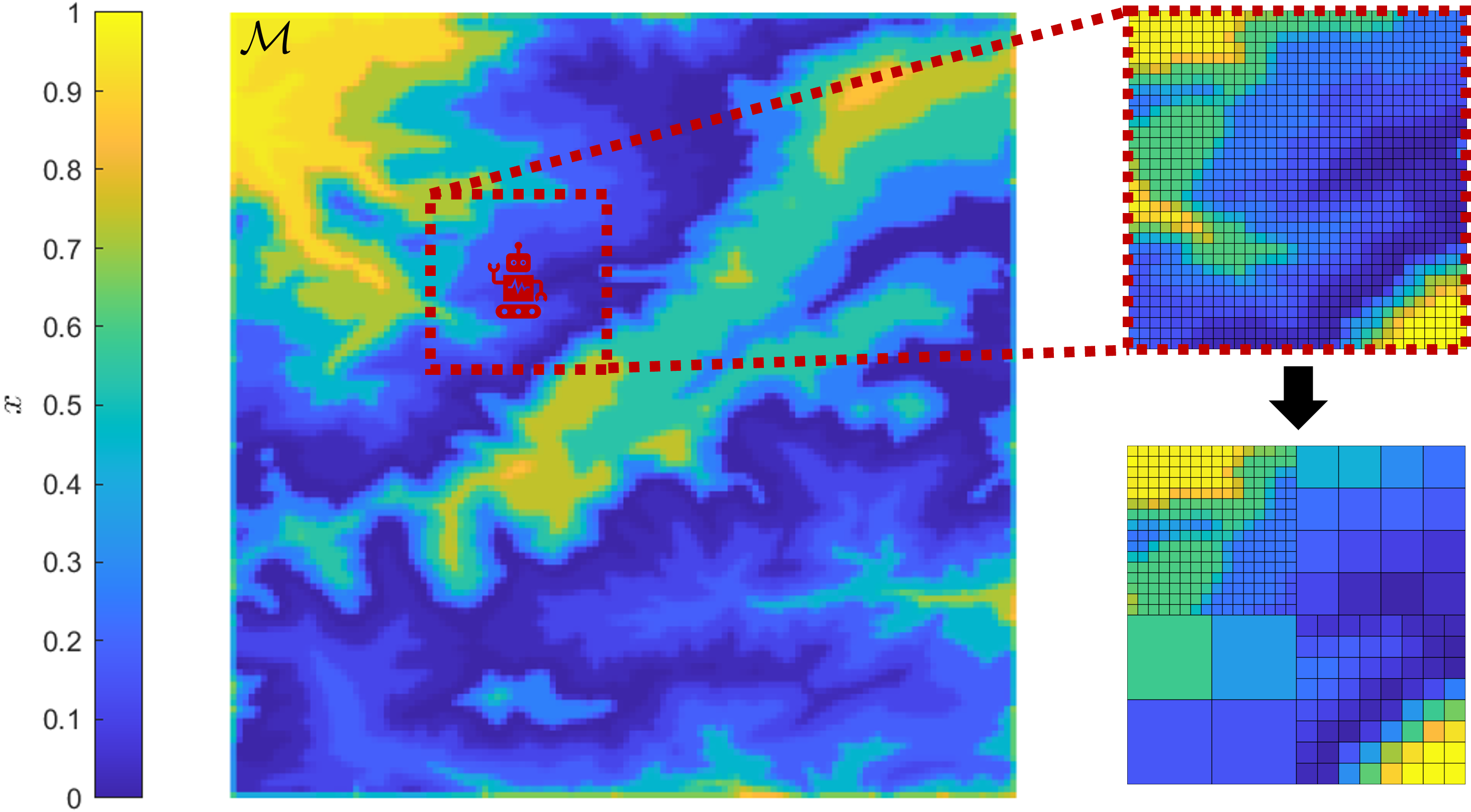}}
    \put(-176,66) {$\leftarrow \hspace{0.02 cm} w \hspace{.02 cm} \rightarrow $}
    \put(-184,72) {\rotatebox{90}{$\leftarrow \hspace{.02 cm} h \hspace{.02cm} \rightarrow $}}
    \caption{Earth traversibility map $\mathcal M$. The mobile robot perceives a local map of dimensions $w \times h$ as it traverses the map and compresses it based on an abstraction template.}
    \vspace{-0.1cm}
    \label{fig:earth_env}
\end{figure}

\textbf{Related Work:}
Several researchers have explored methods for determining what information to communicate based on a given task.
In \cite{Damigos2024}, the authors introduce a communication-aware framework that adjusts data transmission based on communication conditions in centralized multi-agent operations using 3D point cloud maps.
They develop a control function for 5G-enabled multi-robot systems that regulates the transmission rate according to the network's Key Performance Indicators (KPIs).
In \cite{marcotte2020}, the authors propose OCBC, an algorithm that solves a bandid-based combinatorial optimization problem to evaluate an agent's observations.
However, considering the combinatorial nature of the solution, this approach may become computationally intractable with increased robot's field of view.
In \cite{Li2019GraphNN}, the authors present an architecture consisting of a Convolutional Neural Network and a Graph Neural Network to compress and share information between robots for decentralized sequential path-planning.
In \cite{psomiadis2023}, we introduced a communication-aware framework that sequentially selects the optimal map compression to facilitate a robot's path-planning operation on a 2D grid cost map, without requiring the need for neural network training.
In this work, we improve this framework by introducing a new decoder to estimate the unknown map. 
This decoder significantly reduces computational time, allowing for the handling of more realistic scenarios and larger compression template sets.

\textbf{Contributions:}
We propose a communication framework for a mobile sensor robot to assist another robot reach a target within an unknown environment.
The sensor automatically selects and transmits compressed representations of its observed environment in a manner that accounts for both robot's navigation and communication costs.
Our contributions are as follows: 
\begin{enumerate} 
    \item We develop an iterative decoder for environment reconstruction with computational complexity of $O(N^3)$, where $N$ represents the total number of cells in the discretized map.
    \item We account for noise in the robot's measurements and communication channel, and provide a solution that incorporates this uncertainty. 
    \item We perform simulations on large-scale, real-world maps with a moving target. 
\end{enumerate}


\begin{figure}[tb]
    \centering        
    {\includegraphics[width=0.6\linewidth]{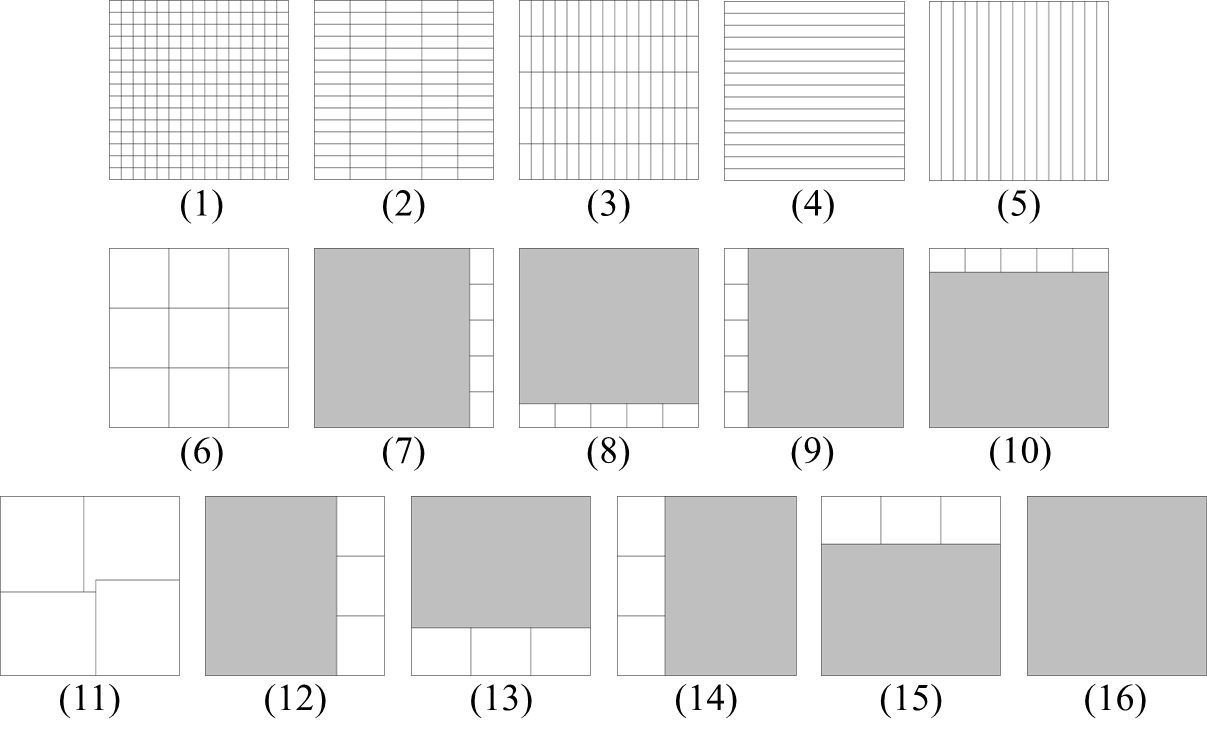}}
    \caption{Example of a set of abstractions. The grey area denotes no information.} 
     \label{fig:abstr_set}
\end{figure}

\section{Preliminaries: Grid World and Abstractions} \label{sec:preliminaries}

We model the environment as a 2D static grid $\mathcal M \subset \Re^2$ (e.g., occupancy grid \cite{elfes1987, moravec1988}, elevation map, or a discretized cost map in general), with the total number of cells in the grid being denoted by $N$.
An example is shown in Figure~\ref{fig:earth_env}.
We define as $M = \{j: \textbf{p}_{j} \in \mathcal{M}\}$ the index set of the finest resolution cells within $\mathcal{M}$, where $\textbf{p}_{j}$ is the position of the $j$-th cell (e.g., cell's center) . 
Each robot can sense a grid of size $w \times h \leq N$, as shown in Figure~\ref{fig:earth_env}. 
We use the vector $x \in [0,1]^{N}$ for the values of the grid cells.  
The $j$-th component of $x$, denoted by $x_j$, lies within the range $[0,1]$ for all $j = 1, \ldots, N$.
$x_j = 1$ indicates a non-traversable cell, while $x_j = 0$ denotes a traversable cell.

We will denote the $j$-th finest resolution cell as $j \in M$.
A robot's measurement on the value of the $j$-th cell is described by $o_j = \mathcal{\alpha}^j x  + v_j$, where $\mathcal{\alpha}^j \in \Re^{1 \times N}$ has 1 at the $j$-th element and zeros everywhere else, and $v \sim \mathcal{N}(0,V)$ is Gaussian noise, due to noisy perception. 
Hence, the sensed grid within each robot's field of view is given by $o = \A x + v$, where $\A = [\mathcal{\alpha}^1; \mathcal{\alpha}^2; \ldots; \mathcal{\alpha}^{w \times h}]$.
Here, ${\A} = {\A}(\textbf{p})$ is a linear mapping ${\A}: [0,1]^{N} \to [0,1]^{w \times h}$, where $\textbf{p}$ is the current robot's position.
For simplicity, the perception noise of certain robots in this study is considered negligible and will be explicitly stated where necessary. 

An \textit{abstraction} is a multi-resolution compressed representation of the grid $\mathcal M$.
Figure~\ref{fig:earth_env} shows an example of an abstracted environment. 
The compressed cells in each abstraction result from applying a \textit{compression template} to the underlying finest resolution grid.
The compression templates are manually crafted beforehand and depend on the form of the robot's sensed map.
Figure \ref{fig:abstr_set} shows an example set of abstraction templates for a $15 \times 15$ grid.
Different techniques have been used to determine the value of a compressed cell \cite{cowlagi2012multiresolution, kraetzschmar2004probabilistic, larsson2020q}.
In this work, the value of a compressed cell is computed by averaging the values of its underlying cells \cite{psomiadis2023}.
Hence, each abstraction ${\A}^{\theta} = {\A}^{\theta}(\textbf{p})$ is a linear mapping ${\A}^{\theta}: [0,1]^{N} \to [0,1]^{k^{\theta}}$, where $k^{\theta}\le wh \le N$ is the number of cells in the compressed grid using abstraction template \({\theta}\). 
If $x \in [0,1]^{N}$ contains the values of the finest resolution map, then the values of the cells using \({\theta}\) are given by $o^{\theta} = \A^{\theta} x$, where $\A^{\theta} \in \Re^{k^{\theta} \times N}$. 
For example, the dimensions of the grid in Figure \ref{fig:abstr_set} are $w = 15$, $h = 15$, and the number of compressed cells for templates $\theta = 4$ and $5$  is $k^{\theta} = 15$, while for $\theta = 6$ it is $k^\theta = 9$, and so forth.
 We will denote a compressed cell as the $i$-th cell.

\begin{figure}[tb]
    \centering        
    {\includegraphics[width=1\linewidth]{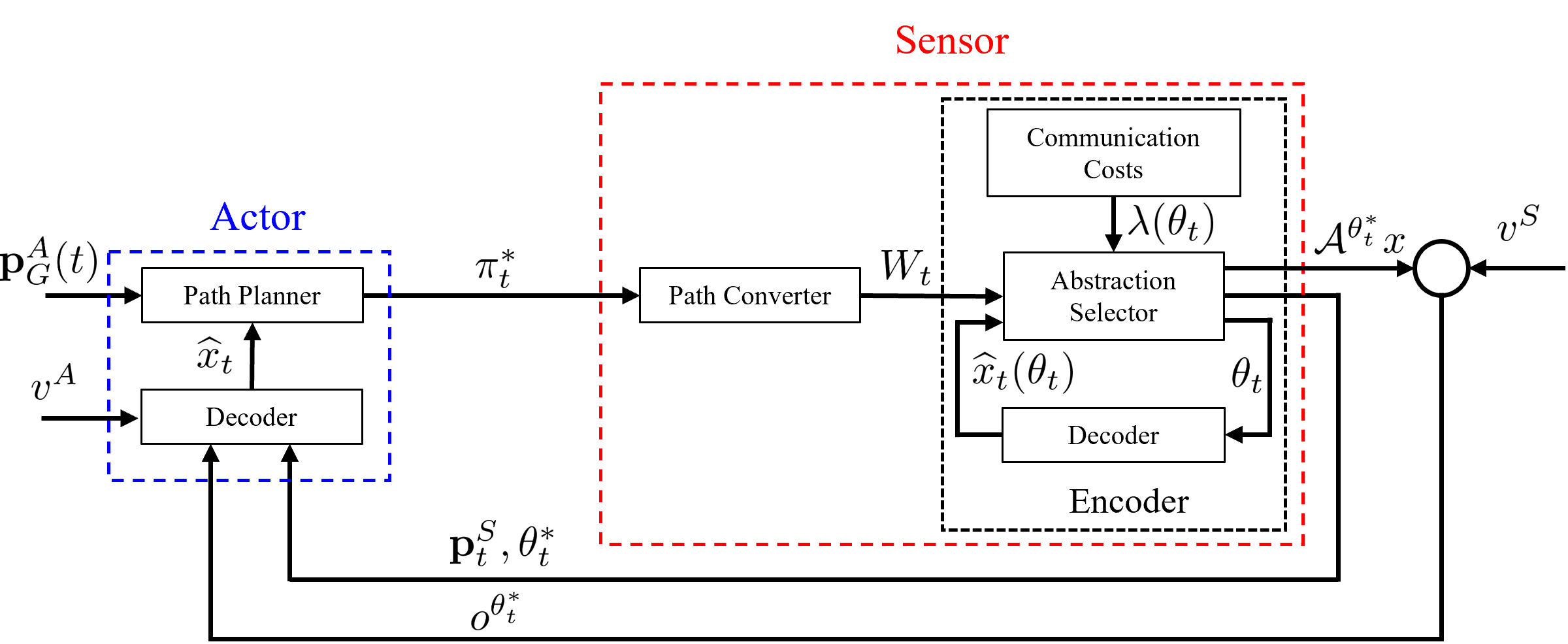}}
    \caption{Proposed framework schematic. The Actor shares its current optimal path \(\pi^*_t\) to the Sensor. The Sensor uses this information to select the optimal abstraction \(\theta^*_t \in \Theta\) of its local map and sends it to the Actor.} 
    \label{fig:framework}
\end{figure}

\subsection{Communication of Abstracted Environments} \label{sec:comm_of_abs}

Let $\Theta$ represent the abstraction set (i.e., codebook) agreed upon by all robots (e.g., see Figure \ref{fig:abstr_set} for a collection of templates). 
In this context, the Actor is the robot that receives the abstraction $\theta$, along with the abstracted local map $o^\theta$, while the Sensor is the one that transmits it. 
At each timestep $t$, the Sensor selects an abstraction $\theta$ and transmits its abstracted environment via a noisy communication channel. 
Specifically, it sends the pair $(o^{\theta_t}, {\theta}_t)$, along with its current position \(\textbf{p}_{t}^S\), where $o^{\theta_t} = \A^{\theta_t} x + v^S$.
Here, $x$ contains the finest resolution measurements sensed by the Sensor, and $v^S \sim \mathcal{N}(0,V^S)$ is Gaussian noise. 
Although the methodology remains unchanged with nonzero Sensor noise, we assume it to be negligible for simplicity and to emphasize the heterogeneity of the robots.
The Actor knows ${\A}^{\theta_t} = {\A}^{\theta_t}(\textbf{p}_t^S)$ and attempts to reconstruct $x$ from $o^{\theta_t}$. 

We compute the total number of bits $n^{\theta}$ required to transmit an abstracted environment as the sum of bits needed to send the measurements $o_i^{\theta_t}$, denoted by $n_m$, and the bits required to transmit an abstraction index ${\theta}$, denoted by $n_{a}$
\begin{equation}\label{eq:bit_abstr}
n^{\theta} = k^{\theta} n_m + n_a.
\end{equation}

\section{Problem Formulation}

Consider a pair of mobile robots, an Actor and a Sensor, that move through an unfamiliar environment \( \mathcal M \subset \Re^2\) with varying traversability conditions.
As they navigate the environment, each robot uses its sensing equipment to observe a portion of the environment (local map).
The Actor is tasked with reaching a specified target while minimizing a certain cost (e.g., path length or time) via employing an online path-planning algorithm. 
Meanwhile, the Sensor aims to assist the Actor in achieving its objective by transmitting informative abstractions of its local map to the Actor at each timestep $t$.
We assume that the Sensor's path is known a priori, and it is an aerial vehicle (i.e., the cell values do not affect its traversability). 
The design of the Sensor's path is not the subject of this work and interested readers may refer to \cite{psomiadis2024multi}.

The positions of the Actor and Sensor at timestep $t$ are denoted by $\textbf{p}_{t}^A, \textbf{p}_{t}^S \in \mathcal M$, respectively. 
Let $u_{t} \in U$ represent the Actor's control action at timestep $t$, chosen from a finite set of control actions $U$.

\subsection{Problem Statement}

In this work, we design a framework where the Sensor facilitates the Actor to reach its target with minimum cost. 
The Sensor selects the optimal abstraction \(\theta^*\) in an online manner, from a given set of abstractions \(\Theta\), to compress its local map in real-time.
The Actor computes its control action $u_t$ based on both its own measurements and the abstractions transmitted by the Sensor at timestep $t$.
The inputs of the framework are the initial positions of the Actor and the Sensor, \(\textbf{p}_{0}^A\) and \(\textbf{p}_{0}^S\), along with the Sensor's predefined path, the Actor's (constant or moving) target \(\textbf{p}_{G}^A(t)\), the set of abstractions $\Theta$, the communication channel noise $v^S$, and the Actor's perception noise $v^A$.
The proposed framework is shown in Figure~\ref{fig:framework}, and the roles of its components are explained in Section \ref{sec:framework_arch}.

\section{Framework Architecture} \label{sec:framework_arch}

In this section, we present the proposed framework, as illustrated in Figure~\ref{fig:framework}. 
The framework incorporates a Path Planner, Path Converter, and Encoder similar to those described in \cite{psomiadis2023}. 
While their functionality is briefly outlined here, any differences from \cite{psomiadis2023} are explicitly highlighted.

\subsection{Decoder} 
\label{sec:Decoder}

The robot's decoder performs the estimation of the map.
Although both robots must use the same estimation strategy for compatibility, each one performs the estimation individually: the Sensor relies solely on its own abstractions, while the Actor uses both its own measurements and the communicated abstractions.
As explained in Section \ref{sec:preliminaries}, both the abstractions and the Actor's measurements can be represented by a set of linear equations $\A_t x + v = o_t$.

There are various techniques to estimate \(x\). 
In \cite{psomiadis2023}, we introduced a decoder that provides an estimate by minimizing the variance of the estimation error, without considering noise in the measurements.
Specifically, the map estimate was the solution of
\begin{subequations}
    \label{eq:decoder_ICRA2024}
    \begin{align}
         \min_{x \in \Re^N} & \| {x} -  \hat{x}_0\|^2, \\
          & A_{0:t} x = o_{0:t}, \label{eq:decod_con_a}\\
          & 0 \leq x_j \leq 1, \textrm{ } j = 1,\ldots,N, \label{eq:decod_con_b}
    \end{align}
\end{subequations}
where $\hat{x}_0$ was the initial belief on the cell values, and $\A_{0:t} = \{\A_0, \A_1, \dots, \A_t\}$ and $o_{0:t} = \{o_0, o_1, \dots, o_t\}$ were the history of abstractions and measurements.

The above problem is a quadratic optimization problem subject to linear constraints. 
State-of-the-art solvers, such as MOSEK 10.1, utilize interior-point methods to address this type of problem. 
Generally, interior-point methods exhibit a computational complexity $O(\sqrt{k}N^3)$ \cite{boyd_convex_optimization}, where $k$ represents the number of constraints and $N$ denotes the number of variables. 
Consequently, solving this optimization problem at each time step $t$ can become a bottleneck over long horizons, particularly when managing a large history of abstractions.

By relaxing the constraint \eqref{eq:decod_con_b}, we can provide an estimation with complexity $O(N^3)$.
It is proven that the minimum variance estimate is equal to the conditional mean \cite{speyer2008}.
Additionally, we can analytically compute the conditional mean $\hat{x}_t'$ and covariance $\Sigma_t$ of the vector containing the values of the map $x$, given measurements or abstractions $o_t = \A_t x + v$, using Kalman Filter,
\begin{subequations} \label{eq:decoder_iterative}
    \begin{align}
    \hat{x}_t' &=\hat{x}_{t-1}' +K_t (o_t - \mathcal{A}_t \hat{x}_{t-1}'),\\
    \Sigma_t &=(I- K_t \mathcal{A}_t) \Sigma_{t-1},\\
    K_t &= \Sigma_{t-1} \mathcal{A}_t ^\top (\mathcal{A}_t \Sigma_{t-1} \mathcal{A}_t ^\top + V)^{-1} \label{eq:decoder_iterative_c},
    \end{align}
\end{subequations}
where $v = v^A  \sim \mathcal{N}(0,V^A)$ for the Actor's perception noise, and $v = v^s  \sim \mathcal{N}(0,V^s)$ for the channel noise.
To ensure that the estimate remains between 0 and 1, we use the estimate projection approach with inequality state constraints \cite{dan2010}.
Hence, the projected estimate is given by:
\begin{equation} \label{eq:projection}
    \hat{x}_t = \max( \min(\hat{x}_{t}', \mathbf{1}), \mathbf{0}),
\end{equation}
where $\textbf{1}$ is a vector of all ones, $\textbf{0}$ is a vector of all zeros, and the operations $\max$ and $\min$ are performed element-wise.

It should be noted that $\hat{x}_t$ does not yield the minimum variance estimate, since this would require the numerical solution of the optimization problem in \eqref{eq:decoder_ICRA2024}, increasing the computational time of the algorithm.
However, the simulations in Section \ref{sec:Sim3} indicate that this relaxation does not impact the final result.

\begin{algorithm}[tb]
\caption{The Actor's Decoder Algorithm}\label{alg:decoder}
\hspace*{\algorithmicindent} \textbf{Input: }{$v^A$, $v^S$, $\hat{x}_{t-1}'$, $\Sigma_{t-1}$, $\theta_{t}$, $o^{\theta_t}$, $\textbf{p}_{t}^A$, $\textbf{p}_{t}^S$} \\
\hspace*{\algorithmicindent} \textbf{Output: } {$\hat{x}_t$, $\hat{x}_t'$, $\Sigma_t$}
\begin{algorithmic}[1]

    \State $(\A^{A}_t,o^{A}_t) \leftarrow $ \textsc{Measure}($L_{t}^A$)

    \State $\hat{x}_t^-, \Sigma_t^- \leftarrow$ \textsc{Decoder}$(\A^{A}_t, o^{A}_t, \hat{x}_{t-1}', \Sigma_{t-1}, v^A)$

    \State $\hat{x}_t', \Sigma_t \leftarrow$ \textsc{Decoder}$(\A^{\theta_t}, o^{\theta_t}, \hat{x}_t^-, \Sigma_t^-, v^S)$

    \State $\hat{x}_t \leftarrow$ \textsc{Projection}$(\hat{x}_t')$ using (\ref{eq:projection})

    \State \Return $\hat{x}_t$, $\hat{x}_t'$, $\Sigma_t$
    
\end{algorithmic}
\end{algorithm}

The decoder's algorithm is presented in Algorithm \ref{alg:decoder}. 
With a slight abuse of notation, we define as $x_0 \sim \mathcal{N}(\hat{x}_0, \Sigma_0)$ the initial Gaussian belief of the map, where $\hat{x}_0$ is the mean and $\Sigma_0$ is the covariance matrix.
The set $L_{t}^A = L_{t}^A(\textbf{p}_{t}^A)$ contains the cells of the Actor's current local map at timestep $t$.
Line 1 in the algorithm employs the function \(\textsc{Measure}\) to sense the values of the cells within the Actor's local map.
Lines 2-3 perform the estimation process using \eqref{eq:decoder_iterative} for the Actor's measurements and the Sensor's abstractions, respectively.
In case the matrix $\mathcal{A}_t^A$ or $\mathcal{A}^{\theta_t}$ exhibits singularities that prevent inversion, we use robust scheme by adding a small amount of artificial noise, $V = \varepsilon I$, to the measurements to mitigate these issues.
Finally, Line 4 uses the function \(\textsc{Projection}\) to project the measurements, so that their values are in $[0,1]$.

The Sensor employs the same estimation algorithm, but omits Lines 1-2 if there is no overlap between the cells it has observed and those the Actor has explored up to timestep $t$. 
When an overlap occurs, the Sensor executes the \(\textsc{Decoder}\) using the overlapping cells, since it knows $v^A$. 
This approach ensures that the Sensor considers the Actor's knowledge of the overlapping cells for the abstraction selection process.
Consequently, if the Actor has already observed a region, the Sensor’s map estimate incorporates this information.

\subsection{Path Planner} \label{sec:path_planner}
The Path Planner employs a graph search algorithm (e.g., Dijkstra's algorithm \cite{Dijkstra1959}), to compute the Actor's path based on the mean value of the estimated map $\hat{x}_t$.
To achieve this, the Path Planner constructs a graph $\mathcal{G} = (\mathcal{N}, \mathcal{E})$ corresponding to the discretized map $\mathcal{M}$, where $\mathcal{N}$ denotes the set of vertices and $\mathcal{E}$ represents the set of edges.
Each vertex in $\mathcal G$ corresponds to the center of a cell in $\mathcal M$. 
We will use \(\textbf{p}\) to denote both the vertex and the cell.
Two vertices are connected by an edge, if the Actor can move between them using a control action \(u \in U\).
The cost of traversing a vertex is given by \cite{psomiadis2023, larsson2021}
\begin{equation}\label{eq:cell_cost}
    c_{t}  (\textbf{p})=     
    \begin{cases}
      \hat{x}_{\textbf{p},t} + a, & \text{if $\textbf{p} \in P_{\epsilon}$},\\
      N(\epsilon + a), & \text{otherwise},
    \end{cases}
\end{equation}
where \(\hat{x}_{\textbf{p},t} \in [0,1]\) is the estimated cell value at position \(\textbf{p}\), \(a\) is a constant penalty for movement, and \(P_{\epsilon} = \{\textbf{p} \in \mathcal{N}: \hat{x}_{\textbf{p},t} \leq \epsilon\}\) is the set of cells meeting a feasibility condition, with the threshold $\epsilon$ satisfying \(\ \hat{x}_{\textbf{p},0} < \epsilon \leq 1, \forall \textbf{p} \in \mathcal N\). 
We require $\epsilon$ to be greater than the initial belief, ensuring that when the Actor must choose between exploring an unknown region or traversing known obstacles, it will select the former.
The intuition behind this cost is based on the assumption that the traversal time of a cell is proportional to its value (difficulty of traversal) and a constant penalty for movement.

\begin{figure}[tb]
     \centering
     \begin{subfigure}[b]{0.28\linewidth}
         \centering
         \includegraphics[width=\textwidth]{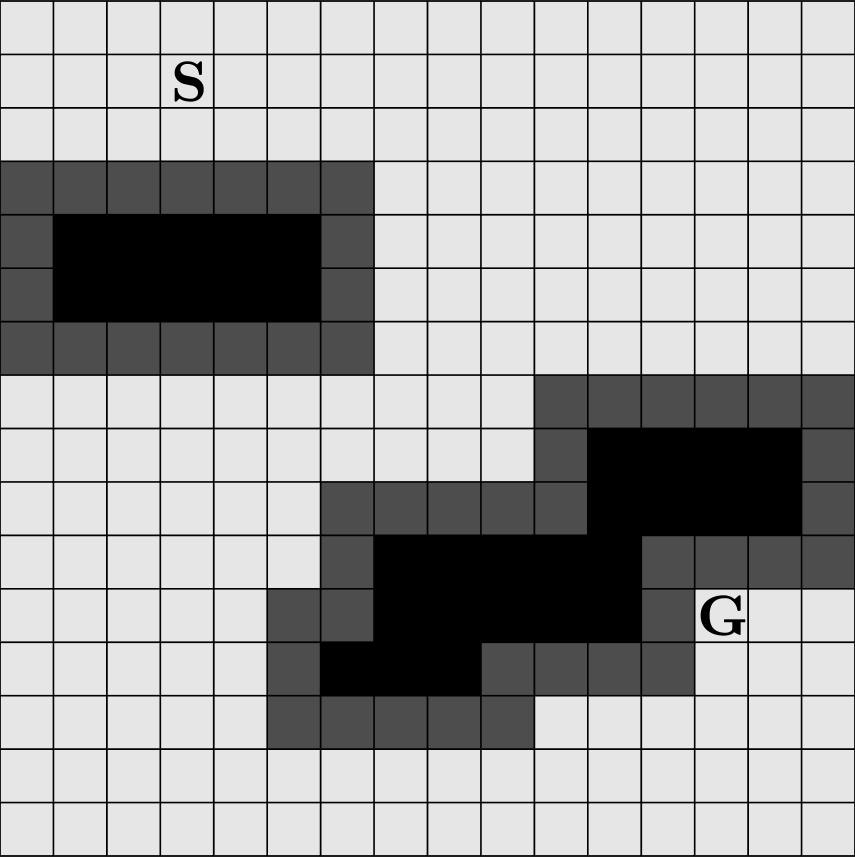}
         \caption{}
         \label{fig:ex_environ}
     \end{subfigure}
     \begin{subfigure}[b]{0.28\linewidth}
         \centering
         \includegraphics[width=\textwidth]{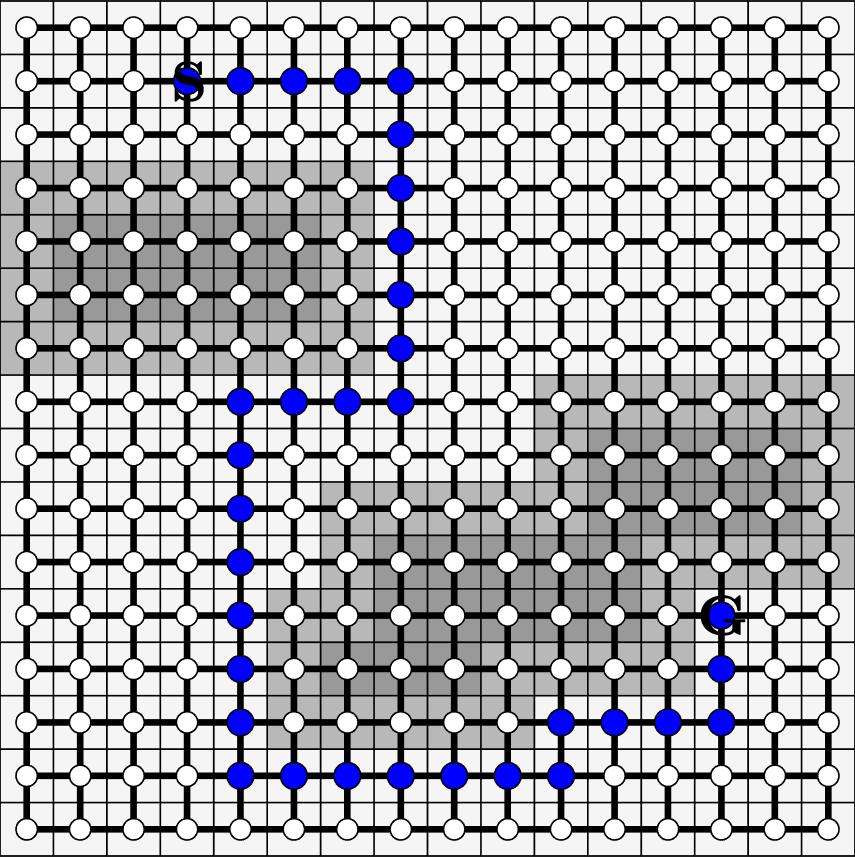}
         \caption{}
         \label{fig:ex_graph}
     \end{subfigure}
     \begin{subfigure}[b]{0.358\linewidth}
         \centering
         \includegraphics[width=\textwidth]{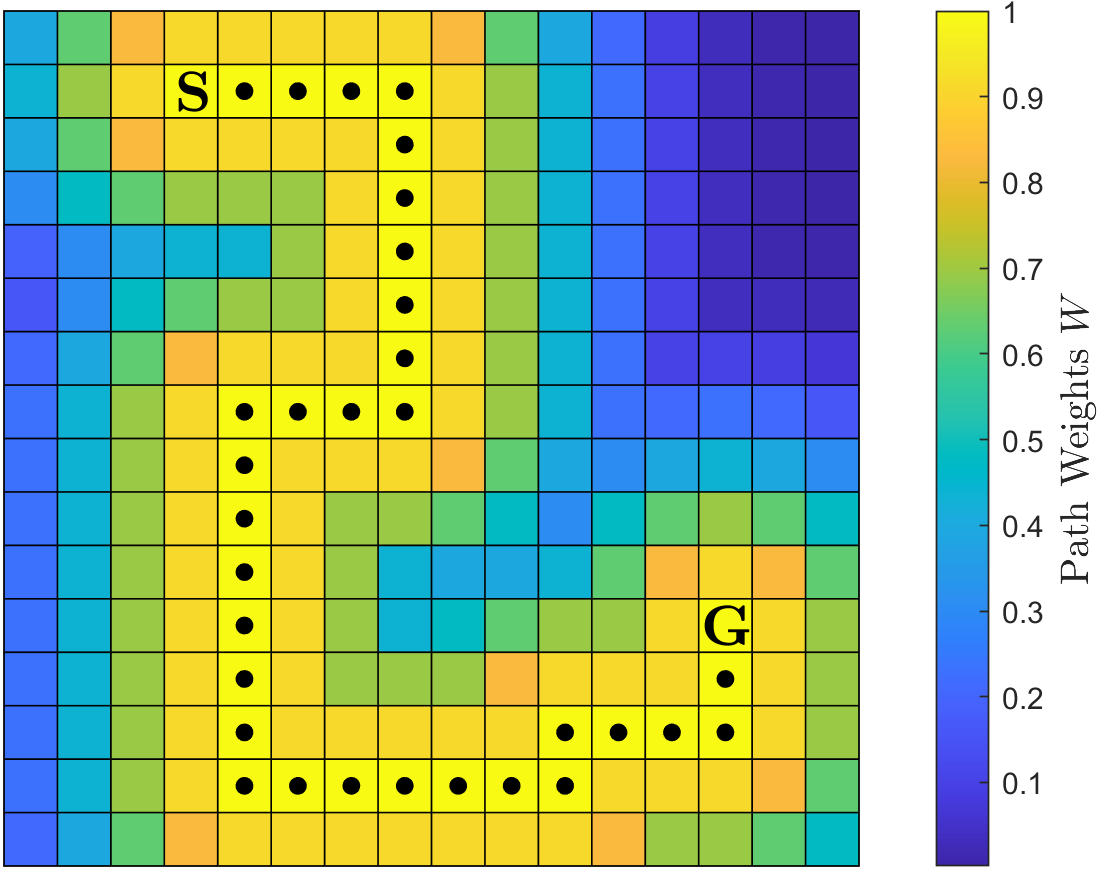}
         \caption{}
         \label{fig:ex_weights}
     \end{subfigure}
        \caption{(a) Discretized environment with static obstacles. \(\textbf{S}\) denotes the starting cell while \(\textbf{G}\) denotes the target; (b) Associated graph of the environment employing the set of actions \(U = \) \{UP, DOWN, LEFT, RIGHT\}, along with the optimal path (blue nodes) with cost function \eqref{eq:cell_cost}; (c) Path weights computed using \eqref{eq:path_weights}.} 
        \label{fig:example}
\end{figure}

Therefore, the Actor's computed path is given by
\begin{equation}\label{eq:path}
    \pi^*_t = \argmin_{\pi \in \Pi_t}\sum_{\textbf{p} \in \pi}{c_{t} (\textbf{p})},
\end{equation}
where \(\Pi_t\) is the set of all paths with the first element being the Actor's current position \(\textbf{p}_{t}^A\), and the last element being its goal location \(\textbf{p}_{G}^A (t)\).
An example of a grid with its associated graph and computed path is presented in 
Figure~\ref{fig:example}\subref{fig:ex_environ} and \ref{fig:example}\subref{fig:ex_graph}.

\subsection{Path Converter}
\label{sec:path_converter}

The Sensor's Path Converter identifies a region of interest around the Actor's current path to guide the abstraction selection process. 
After receiving the Actor's current path $\pi^*_t$, the Path Converter assigns weights to each cell in $\mathcal M$ based on its proximity to the path.
The weights are given by
\begin{equation}\label{eq:path_weights}
    w_t(\textbf{p}) = \max_{\textbf{p}^* \in \pi^*_t} e^{-\frac{\|\textbf{p}-\textbf{p}^*\|^2}{2\sigma^2}}, \qquad \textbf{p} \in \mathcal M,
\end{equation}
where \(\sigma\) is a scalar characterizing the width around the path. 
Figure \ref{fig:example}(\subref{fig:ex_weights}) presents an example of path weights.

\begin{figure*}[!t]
    \centering   
     \begin{subfigure}[b]{0.23\linewidth}
         \centering
         \includegraphics[width=\textwidth]{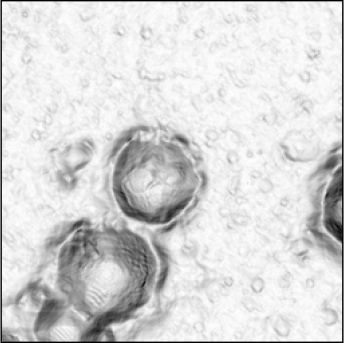}
         \caption{Full Map}
         \label{fig:env_mars}
     \end{subfigure}
     \hspace{1mm}
     \begin{subfigure}[b]{0.23\linewidth}
         \centering
         \includegraphics[width=\textwidth]{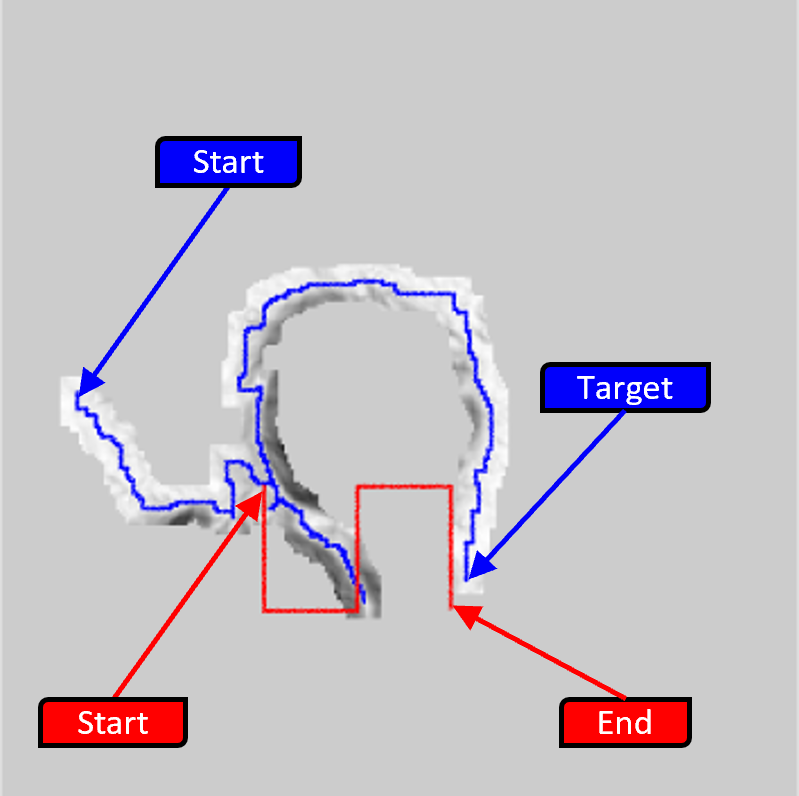}
         \caption{U Framework}
         \label{fig:U_final_scen1}
     \end{subfigure}
     \hspace{1mm}
     \begin{subfigure}[b]{0.23\linewidth}
         \centering
         \includegraphics[width=\textwidth]{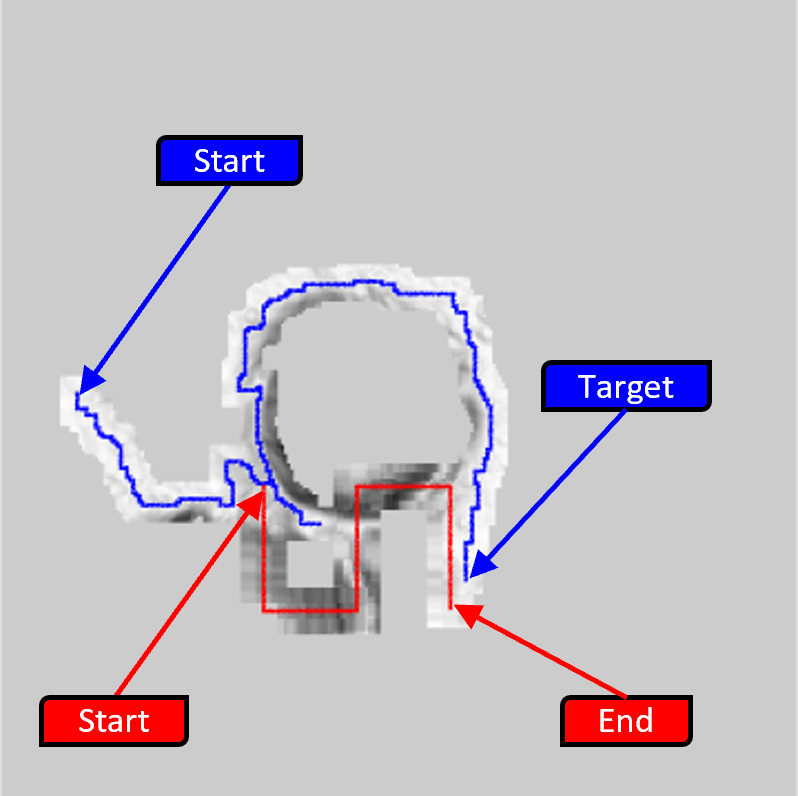}
         \caption{AS Framework}
         \label{fig:AS_final_scen1}
     \end{subfigure}
     \hspace{1mm}
     \begin{subfigure}[b]{0.23\linewidth}
         \centering
         \includegraphics[width=\textwidth]{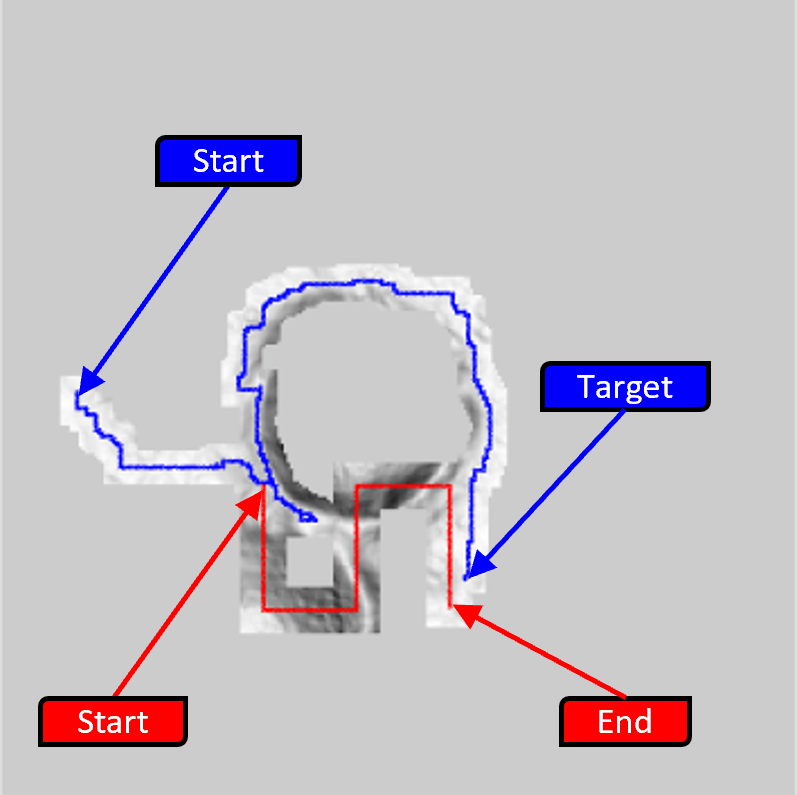}
         \caption{FI Framework}
         \label{fig:FI_final_scen1}
     \end{subfigure}

    \caption{(a) Mars inclination map computed from the depth map in \cite{HiRISE2024}; (b-d) Final estimated map with $v^A  \sim \mathcal{N}(0,10^{-5}I)$ for (b) Uninformed Framework, (c) Abstraction-Selection Framework, (d) Fully-Informed Framework. Blue and red indicate the cells that the Actor and the Sensor have already traversed, respectively.}
    \label{fig:mars_fig}
    \vspace{-0.6cm}
\end{figure*}

\subsection{Encoder}
\label{sec:encoder}

The Sensor's Encoder selects the optimal abstraction to communicate to the Actor. 
Let $\theta_t \in \Theta$ represent the Sensor's abstraction at timestep $t$, where $\Theta$ denotes the set of available abstractions. 
Unlike the Encoder in \cite{psomiadis2023}, this Encoder does not rely on the history of abstractions to compute the next abstraction, as all the necessary past information is contained in $\hat{x}_t$. 
Hence, the optimal abstraction at time $t$ is determined through an exhaustive search using the following function
\begin{subequations}\label{eq:encoder}
    \begin{eqnarray}
         & J_t(\theta_{t}) = \| W_t \circ (\tilde{x}_t-\hat{x}_t(\theta_{t}))\|^2 + \lambda(\theta_t), \label{eq:encoder_a}\\
         & \theta^*_t = \argmin_{\theta_t \in \Theta} J_t(\theta_{t}), \label{eq:encoder_b}
    \end{eqnarray}
\end{subequations}
where \(W_t \in \Re^N\) is the vector containing the weights \(w_t(\textbf{p})\) of every cell, \(\tilde{x}_t \in [0,1]^N\) is the Sensor's sensed cell values until time \(t\), \(\hat{x}_t(\theta_{t}) \in [0,1]^N\) is the estimation vector computed from \eqref{eq:decoder_iterative}, and \(\lambda(\theta_{t})\) is the communication cost (related to $n^\theta$ in \eqref{eq:bit_abstr}). 
Note that in our case, the Sensor's perception noise is assumed to be negligible, so $\tilde{x}_t$ contains the true values of the cells the Sensor has already observed. 
To compute \(\hat{x}_t\) for each abstraction $\theta_{t}$, we assume that the selection is based on the most likely noise realization, $v = 0$, which leads to the observation $o^{\theta_t} = \A^{\theta_t} x$.
Note that the Encoder's computational cost is proportional to the cost of the Decoder and the number of the abstractions in the set.
The Encoder's algorithm is similar to the one presented in \cite{psomiadis2023}, with the preceding differences taken into account.

\section{Simulations}
For the simulations, we utilized two distinct satellite maps, shown in Figure~\ref{fig:earth_env} and Figure~\ref{fig:mars_fig}(\subref{fig:env_mars}). 
Figure~\ref{fig:earth_env} depicts an $128 \times 128$ Earth traversability map.
Figure~\ref{fig:mars_fig}(\subref{fig:env_mars}) depicts a $256 \times 256$ inclination map of Mars. 
This map is derived from the depth data provided in \cite{HiRISE2024}. 
To apply our framework, we first need to normalize the map.
Let $y_j$ represent the value of the $j$-th cell in the depth map. 
The inclination value for the $j$-th cell, $x_j$, is computed by $x_j = \frac{z_j - \min{z}} {\max{z} - \min{z}}$, where $z_j = \sum_{j' \in J}{|y_j - y_j'|}$, and $J$ denotes the index set of neighboring cells to the $j$-th cell.

Three different scenarios were studied using these maps.
In the first scenario, the Mars map was used to evaluate our algorithm in a challenging environment, characterized by a large map size and varying levels of Actor perception noise.
The second scenario is a target tracking operation, where the objective is for the Actor to reach a moving target with a known location on the Earth's map.
In the third scenario, we compared the performance of our proposed framework with that of the framework presented in \cite{psomiadis2023} in terms of computational time.
In all scenarios, we utilized a single Actor-Sensor pair, both starting simultaneously and moving one cell per timestep. 
The Sensor, designed to operate as a surveillance drone, has the capability to move over obstacles and follows a predefined lawnmower path, as shown in Figure~\ref{fig:mars_fig}.
Both robots are positioned at the center of their respective local maps and the Actor's control set is \(U = \)\{UP, DOWN, LEFT, RIGHT\}.

The Actor's Path Planner parameter in \eqref{eq:cell_cost} is \(a = 0.025\), which was tuned based on the range of values of $\hat{x}$. $\epsilon$ varies depending on the specific scenario, as detailed in Section \ref{sec:path_planner}.
For the parameter in \eqref{eq:path_weights}, we use \(\sigma = 20\).
The communication cost, as defined in \eqref{eq:encoder}, is given by $\lambda (\theta) = 0.02 k^\theta$, where $k^\theta$ depends on the resolution of the abstraction template, indicating that the cost of abstraction is proportional to the cardinality of the abstracted map.

\vspace{-0.1cm}
\subsection{Performance Metrics}

In the first two scenarios, we compare our Abstraction Selection (AS) framework to a Fully-Informed (FI) framework, where in the latter the Sensor transmits all observed cells at each timestep. 
The comparison focuses on their accumulated path cost and communication cost.
In the third scenario, we compare the proposed framework to the one presented in \cite{psomiadis2023} based on the average computational time of the Actor's Decoder using a computer equipped with a 2.3 GHz, 8-Core, 11th Gen Intel Core i7-11800H CPU and 16GB of RAM. 
The simulations were conducted in MATLAB.

We compute the Actor's accumulated cost using $\mathcal{C} = \sum_{\textbf{p} \in \pi_f}{c (\textbf{p})}$, where ${c (\textbf{p})}$ is computed using the first scale of \eqref{eq:cell_cost}, and \(\pi_f\) represents the set of all cells the Actor traversed to reach its goal, including any duplicate cells.
The average cost ratio is
\begin{equation}\label{eq:time_ratio}
    r_{\textrm{cost}} = \frac{1}{n_{\textrm{sim}}}\sum_{i = 0}^{n_{\textrm{sim}}}\frac{\mathcal{C}(i)}{\mathcal{C}_{\textrm{max}}(i)},
\end{equation}
where \(n_\textrm{sim}\) is the total simulation number, \(\mathcal{C}(i)\) denotes the accumulated cost for simulation $i$, and $\mathcal{C}_{\textrm{max}}$ is the accumulated cost employing an Uninformed (U) framework, where the Actor reaches its target without any assistance from the Sensor.
For the communication cost, we compute the average ratio of the transmitted bits
\begin{equation}\label{eq:bit_ratio}
    r_{\textrm{bits}} = \frac{1}{n_{\textrm{sim}}} \sum_{i = 0}^{n_{\textrm{sim}}} \frac{\mathcal{B}(i)}{\mathcal{B}_{\textrm{max}}(i)},
\end{equation}
where $\mathcal{B} = \sum_{t = 0}^{T^S}{n_{t}}$ with \(n_{t}\) given in (\ref{eq:bit_abstr}) and referring to the number of bits sent by the Sensor at \(t\), and $T^S$ is the Sensor's time horizon (equal to the number of timesteps that it operates). 
We set \(n_m = 12\), and \(n_a = 4\).
$\mathcal B_{\textrm{max}}$ refers to the communication cost $\mathcal{B}$ of the FI framework.

\subsection{Scenario 1: Mars}

The objective of this scenario is for the Actor to reach a stationary location on the Mars map, shown in Figure~\ref{fig:mars_fig}(\subref{fig:env_mars}).
The Actor's field of view is an \(11 \times 11\) grid centered on its current position, its initial position is \(\textbf{p}_{0}^A = (25,130)\) and aims to reach its goal at \( \textbf{p}_{G}^A = (150,70)\). 
The Sensor covers a larger area with a \(15 \times 15\) grid, and uses a finite set of 16 abstractions, depicted in Figure~\ref{fig:abstr_set}.
It starts at \(\textbf{p}_{0}^S = (85,100)\), and its 
time horizon is $T^S = 180$.
The communication channel noise is modelled as $v^S  \sim \mathcal{N}(0,10^{-4}I)$.
The threshold parameter of the Path Planner is set to $\epsilon = 0.201$, chosen to ensure the robot avoids traversing areas with inclinations beyond a certain limit for safety and power-saving reasons.
In our map, this value corresponds to $2.2m$ per cell of average inclination.
The initial belief is set to $x_0 \sim \mathcal{N}(0.2, I)$ since its mean needs to be less than $\epsilon$.

\begin{table}[!t]
\caption{Results for Scenario 1 (Mars).}
\label{tab: results1}
\centering
\small
\begin{tabular}{ |c|c|c|c|c|c| } 
\hline
& \multicolumn{2}{|c|}{$v^A  \sim \mathcal{N}(0,10^{-3}I)$} & \multicolumn{2}{|c|}{$v^A  \sim \mathcal{N}(0,10^{-5}I)$} \\

    \hline
    {Metrics} &  {AS Fram.} &  {FI Fram.} &  {AS Fram.} &  {FI Fram.}  \\
    \hline
    ${\mathcal{C}(i)}$
    & 55.9 & 47.3  & 52.2  &  47.5 \\
    \hline
    ${\mathcal{B}(i)}$
     & 7692  & 486000  & 7680  &  486000 \\
    \hline
    \(r_\textrm{cost}\) (\%) & 55.3 & 46.8 & 77.6  & 70.7 \\ 
    \hline
    \(r_\textrm{bits}\) (\%) & 1.6 & 100 & 1.6 & 100 \\ 
    \hline
\end{tabular}
\end{table}

The results for two different values of Actor's perception noise $v^A$ are presented in Table \ref{tab: results1}.
Considering that $r_{\textrm{cost}}$ is computed using the U framework, we can conclude that both the FI and AS frameworks significantly reduced the accumulated cost compared to the U Framework. 
However, our proposed AS framework stands out by achieving a remarkable 98\% reduction in communicated information compared to the FI framework.
It is important to note that, in the FI framework, the Sensor transmits all observed data at each timestep due to the presence of noise. 
If the noise was negligible, the Sensor would only need to send the newly observed cells.
Figure~\ref{fig:mars_fig}(\subref{fig:U_final_scen1} - \subref{fig:FI_final_scen1}) presents the final timestep of the simulation by using the three frameworks.

\subsection{Scenario 2: Earth (Moving Target)}
In this scenario, we conducted 100 simulations using the Earth map shown in Figure~\ref{fig:earth_env}. 
In each simulation, the target starts at position \( \textbf{p}_{G}^A(t = 0) = (90,49)\) and follows a random path. 
To ensure that the Actor can always reach the target, we assume that the target moves every even timestep.
The Sensor's field of view and set of abstractions is the same as in scenario 1, its initial position is $\textbf{p}_{0}^S = (46, 62)$, and its time horizon is $T^S = 105$.
The Actor's filed of view is $5 \times 5$ and its initial location is $\textbf{p}_{0}^A = (12,57)$.
The characteristics of the noise are 
$v^S  \sim \mathcal{N}(0,10^{-5}I)$, and $v^A  \sim \mathcal{N}(0,10^{-6}I)$.
The Path Planner parameter is $\epsilon = 0.501$, while the map's initial belief is $x_0 \sim \mathcal{N}(0.5, I)$.

\begin{table}[!t]
\caption{Results for Scenario 2 (Earth).}
\label{tab: results2}
\centering
\small
\begin{tabular}{ |c|c|c|c| } 
    \hline
    {Metrics} &  {AS Fram.} &  {FI Fram.}  \\
    \hline
    $\mu_{\mathcal{C}} \pm \sigma_{\mathcal{C}}$
    & 19.4 $\pm$ 0.9 & 16.7 $\pm$ 6.5 \\
    \hline
    $\mu_{\mathcal{B}} \pm \sigma_{\mathcal{B}}$
     & 6602 $\pm$ 115 & 283500 $\pm$ 0 \\
    \hline
    \(r_\textrm{cost}\) (\%) & 22.2 & 19.0  \\ 
    \hline
    \(r_\textrm{bits}\) (\%) & 2.3 & 100 \\ 
    \hline
\end{tabular}
\end{table}

Table \ref{tab: results2} presents the results of the simulations. 
The first two metrics are the mean and the standard deviation of the results over all the 100 simulations.
We observe that the AS framework performs similarly to the FI framework, based on their accumulated cost. 
Additionally, similar to scenario 1, the AS Framework achieves a 98\% reduction in communicated information, compared to the FI framework.

\subsection{Computational Time Analysis}
\label{sec:Sim3}

For the third scenario, we compared our proposed framework with the one introduced in \cite{psomiadis2023}, based on the computational time of the Actor's Decoder Algorithm (see Algorithm \ref{alg:decoder}).
To assess the algorithm's complexity, we varied the Sensor's time horizon $T^S$, as it is proportional to the number of sent abstractions. 
We used the map shown in Figure~\ref{fig:earth_env}.

The Actor has the same field of view and initial position as in scenario 2, while its target location is fixed at $\textbf{p}_{G}^A = (86, 49)$.
The Sensor's field of view is $7 \times 7$, its start location is $\textbf{p}_{0}^S = (46, 62)$, and utilizes a finite set of 10 abstractions, as illustrated in Figure 6 of \cite{psomiadis2023}.
No noise is introduced in the simulations, as \cite{psomiadis2023} does not provide a straightforward method for handling noise. 
The initial belief and the Path Planner parameter are the same as in scenario 2.

\begin{figure}[t]
    \centering        
    {\includegraphics[width=0.8\linewidth]{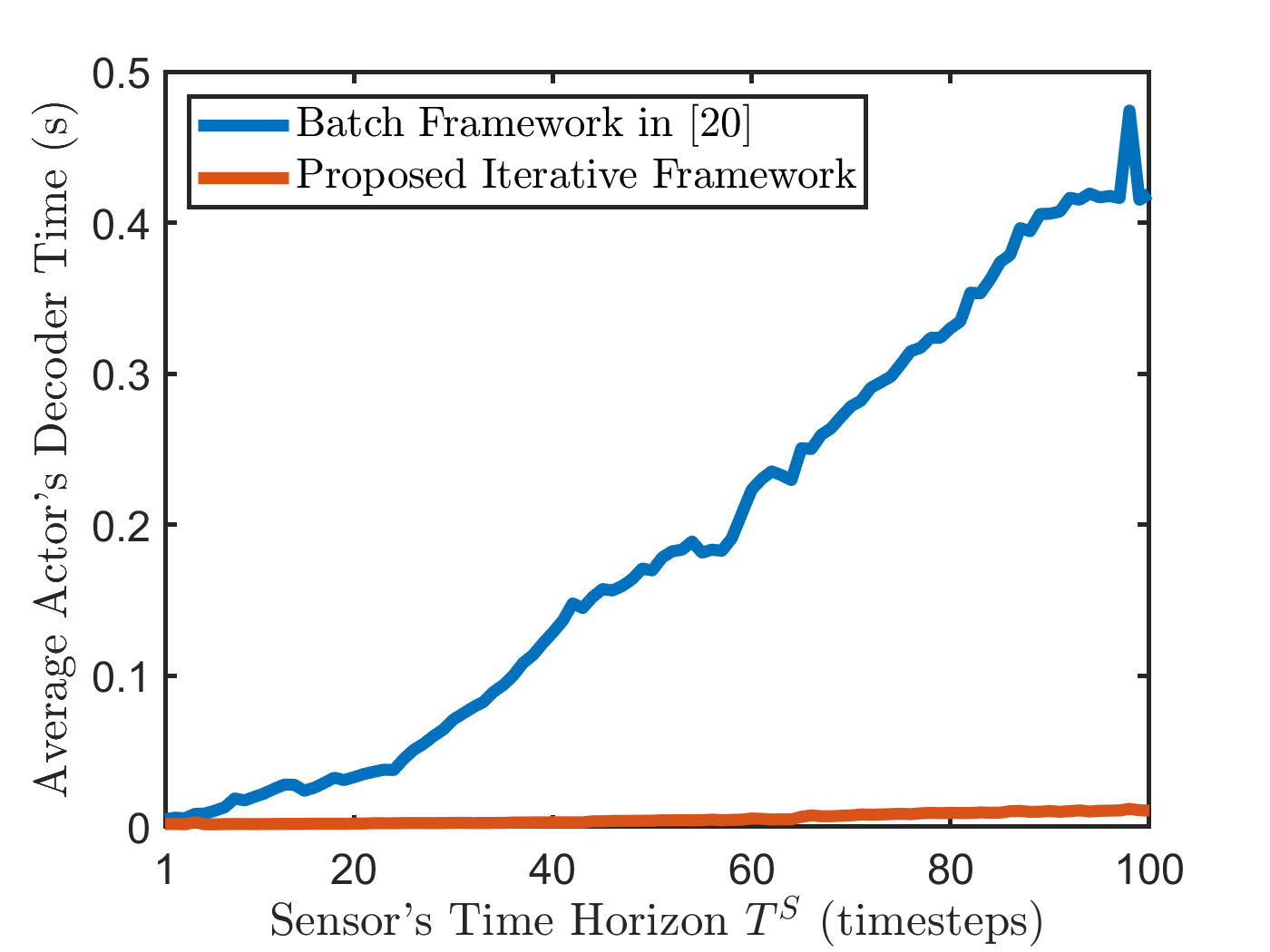}}
    \caption{Average computational time of Actor's Decoder vs. Sensor's time horizon $T^S$.}
    \label{fig:sim3}
\end{figure}

Figure~\ref{fig:sim3} shows the average Actor's Decoder runtime (over all timesteps of each simulation) for 100 simulations, where each simulation is performed with a different Sensor Time Horizon $T^S$.
We observe that the rate of increase in time using the framework from \cite{psomiadis2023} is significantly larger than with the proposed framework. 
This is because the former relies on the history of abstractions, considering that the Sensor's time horizon $T^S$ is proportional to the number of transmitted abstractions by the end of the simulation.
It should be noted that although the Decoder in \cite{psomiadis2023} provides slightly better estimation compared to the proposed Decoder, it ultimately does not impact the Actor's accumulated cost $\mathcal{C}$. 
Therefore, its effect on the problem's objective is negligible.

\section{Conclusions}

In this paper, we addressed the problem of facilitating a robot's path-planning operation using a mobile sensor. 
The sensor selects the optimal compressed representation of its local map, transmitting it to the robot while automatically considering both communication costs and the predicted robot trajectory.
We proposed a novel framework with a decoder that estimates the unknown environment with reduced computational complexity compared to the state-of-the-art approach, while also handling noise. 
Our simulations demonstrated the framework's effectiveness, achieving a 98\% reduction in transmitted data compared to sending raw data, while maintaining comparable path lengths and low computational time. 
Future work will focus on implementing this framework in real-world UAV-UGV collaborative operations.




\bibliographystyle{IEEEtran}
\bibliography{IEEEabrv,refs,Maity,Pedram}

\end{document}